\documentclass[conference]{IEEEtran}
\IEEEoverridecommandlockouts
\usepackage[table]{xcolor}
\usepackage{graphicx}
\usepackage{float}
\usepackage{verbatim}
\usepackage{xcolor}
\usepackage{multirow}
\usepackage{booktabs} 
\usepackage{stackengine}
\usepackage{amsfonts}

\usepackage{amsmath}
\usepackage{wasysym}
\usepackage{colortbl}

\usepackage{graphicx}

\usepackage[mathscr]{eucal}
\usepackage{epsfig,epsf,psfrag}
\usepackage[caption=false]{subfig} 
\usepackage{fixltx2e}
\usepackage{array}
\usepackage{verbatim}
\usepackage{bm}
\usepackage{verbatim}
\usepackage{textcomp}
\usepackage{mathrsfs}
\usepackage{epstopdf}

\catcode`~=11 \def\UrlSpecials{\do\~{\kern -.15em\lower .7ex\hbox{~}\kern .04em}} \catcode`~=13 

\allowdisplaybreaks[3]

\newcommand{\calL}{\mathcal{L}}

\newcommand{\rmf}{\mathrm{f}}

\newcommand{\frakL}{\mathfrak{L}}

\DeclareMathAlphabet{\mathbsf}{OT1}{cmss}{bx}{n}
\DeclareMathAlphabet{\mathssf}{OT1}{cmss}{m}{sl}

\DeclareSymbolFont{bsfletters}{OT1}{cmss}{bx}{n}  
\DeclareSymbolFont{ssfletters}{OT1}{cmss}{m}{n}
\DeclareMathSymbol{\bsfGamma}{0}{bsfletters}{'000}
\DeclareMathSymbol{\ssfGamma}{0}{ssfletters}{'000}
\DeclareMathSymbol{\bsfDelta}{0}{bsfletters}{'001}
\DeclareMathSymbol{\ssfDelta}{0}{ssfletters}{'001}
\DeclareMathSymbol{\bsfTheta}{0}{bsfletters}{'002}
\DeclareMathSymbol{\ssfTheta}{0}{ssfletters}{'002}
\DeclareMathSymbol{\bsfLambda}{0}{bsfletters}{'003}
\DeclareMathSymbol{\ssfLambda}{0}{ssfletters}{'003}
\DeclareMathSymbol{\bsfXi}{0}{bsfletters}{'004}
\DeclareMathSymbol{\ssfXi}{0}{ssfletters}{'004}
\DeclareMathSymbol{\bsfPi}{0}{bsfletters}{'005}
\DeclareMathSymbol{\ssfPi}{0}{ssfletters}{'005}
\DeclareMathSymbol{\bsfSigma}{0}{bsfletters}{'006}
\DeclareMathSymbol{\ssfSigma}{0}{ssfletters}{'006}
\DeclareMathSymbol{\bsfUpsilon}{0}{bsfletters}{'007}
\DeclareMathSymbol{\ssfUpsilon}{0}{ssfletters}{'007}
\DeclareMathSymbol{\bsfPhi}{0}{bsfletters}{'010}
\DeclareMathSymbol{\ssfPhi}{0}{ssfletters}{'010}
\DeclareMathSymbol{\bsfPsi}{0}{bsfletters}{'011}
\DeclareMathSymbol{\ssfPsi}{0}{ssfletters}{'011}
\DeclareMathSymbol{\bsfOmega}{0}{bsfletters}{'012}
\DeclareMathSymbol{\ssfOmega}{0}{ssfletters}{'012}

\newcommand{\tilw}{\tilde{w}}

\newtheorem{data model}{Data Model}

\newcommand{\qednew}{\nobreak \ifvmode \relax \else
      \ifdim\lastskip<1.5em \hskip-\lastskip
      \hskip1.5em plus0em minus0.5em \fi \nobreak
      \vrule height0.75em width0.5em depth0.25em\fi}

\newcommand{\cmmnt}[1]{}

\title{Entropy-Based Modeling for Estimating Soft Errors Impact on Binarized Neural Network Inference}

\author{Navid Khoshavi$^{1, 2}$, Saman Sargolzaei$^1$, Arman Roohi$^3$, 
Connor Broyles$^2$, Yu Bi$^4$ 
\\$^1$Department of Computer Science, Florida Polytechnic University
\\ $^2$Department of Electrical and Computer Engineering, Florida Polytechnic University
\\$^3$Department of Electrical and Computer Engineering, University of Texas, Austin
\\$^4$Department of Electrical and Computer Engineering, University of Rhode Island}

\date{}

\begin{document}

\maketitle

\begin{abstract}
Over past years, the easy accessibility to the large scale datasets has significantly shifted the paradigm for developing highly accurate prediction models that are driven from Neural Network (NN). These models can be potentially impacted by the radiation-induced transient faults that might lead to the gradual downgrade of the long-running expected NN inference accelerator. The crucial observation from our rigorous vulnerability assessment on the NN inference accelerator demonstrates that the weights and activation functions are unevenly susceptible to both single-event upset (SEU) and multi-bit upset (MBU), especially in the first five layers of our selected convolution neural network. In this paper, we present the relatively-accurate statistical models to delineate the impact of both undertaken SEU and MBU across layers and per each layer of the selected NN. These models can be used for evaluating the error-resiliency magnitude of NN topology before adopting them in the safety-critical applications.
\end{abstract}

\begin{IEEEkeywords} 
\textit{Fault Injection, Deep Neural Network Accelerator, Machine Learning, Soft Error, Statistical Model}
\end{IEEEkeywords} 


\section{Introduction}
\label{sec:introduction}
Over the past few decades, the focus of the researchers has been on speeding up the computational capabilities in the traditional computing-centric model, which is based on moving large volumes of data from/to memory storage to/from processing nodes for execution/store. To maximize the efficacy of computing-centric models, many fine/coarse -grain parallelism paradigms have been utilized, from a software approach, i.e. instruction/data/thread/transaction -level parallelism, to micro-architecture domain, i.e. pipelining \cite{murakami1989simp}, superscalar \cite{palacharla1997complexity}, VLIW, Single Instruction Multiple Data (SIMD) instructions \cite{asanovic1992spert}, vector processors \cite{ciricescu2003reconfigurable}, Graphics Processor Units (GPUs), and chip-multiprocessors (CMPs).

While the aforementioned innovations have resulted in significant energy reduction per Floating-Point Operation (FLOP), the energy cost of data movement has marginally reduced via adopting these approaches. This has encouraged the computer architecture designers to explore shifting the paradigm of computation from the computing-centric models towards the fine-tuned application-specific architectures for performing a specific task with the highest performance while incurring the minimum energy cost per computation operation. Among the applications that have received considerable attention for architectural customization, the hardware accelerators for Machine Learning (ML) and Deep Learning (DL) algorithms are the most well-studied designs. To optimize the architecture design for ML/DL accelerators, a broad spectrum of methodologies have been proposed at both software and hardware levels. \cmmnt{The data-layout compression \cite{lin2018supporting}, network pruning, encoding, batch-size reduction, \cite{parashar2017scnn}, and reduced precision representation \cite{de2017understanding} are among the techniques that have been widely adopted and explored in both academia and industry sectors.}

\begin{figure}[t!]
\centering
    \includegraphics[width=0.45\textwidth]{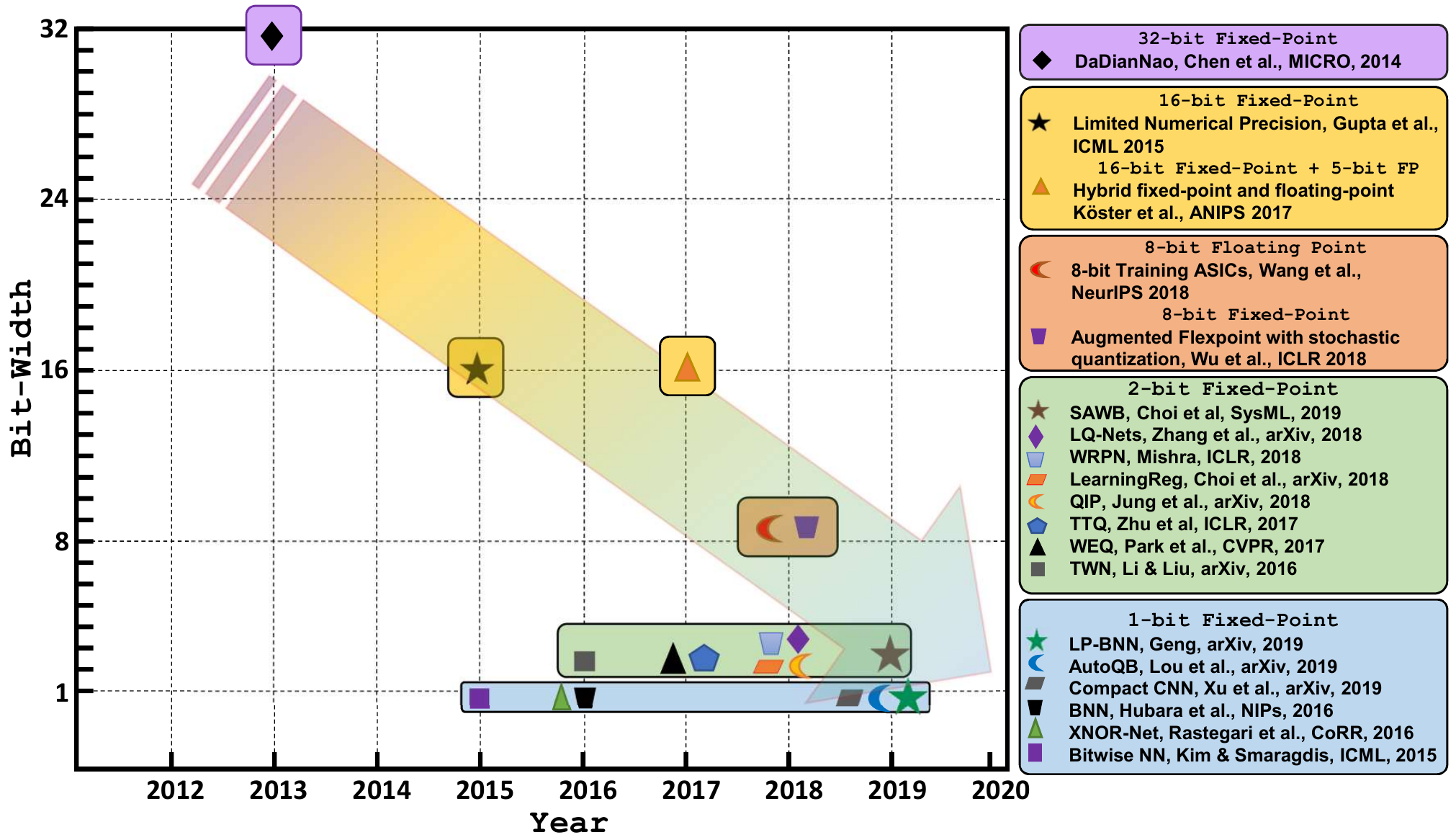}
  \vspace{-10pt}
  \caption{Roadmap of reduced bit-width representation in neural networks.}
    \label{fig:BNN_trend}
\end{figure}

In the algorithm-based approaches, the data-layout compression \cite{Sedghi2017SIE,sedghi2017learning,lin2018supporting}, network pruning, encoding, batch-size reduction \cite{parashar2017scnn}, subset selection \cite{Sedghi2020rep}, and reduced precision representation \cite{de2017understanding} have been explored extensively in both academia and industry sectors. In particular, the employment of weights and activations with low bit-width decreases the model size and the computing complexity. For instance, in \cite{chen2014dadiannao}, the floating-point network parameters quantized to 32-bit fixed-point, which leads to a reduction in both data movement overhead and computation cost. Further improvement performed in \cite{gupta2015deep}, where authors leveraged a 16-bit fix-point representation for weights. In \cite{koster2017flexpoint}, hybrid representations, including both fixed and floating-point, were utilized, where some layers were quantized, while the rest have remained unchanged. This trend, reduction in precision, has been aggressively continued to compress the representative bits to 8-bit fixed-point fashions \cite{wu2018training, wang2018training}, 2-bit fixed-point precision \cite{li2016ternary, park2017weighted, zhu2016trained, jung2018joint, choi2018learning, mishra2017wrpn, zhang2018lq, wu2018training}, and 1-bit fixed-point representation or Binarized Neural Network (BNN) \cite{hubara2016binarized, ar219, kim2016bitwise, rastegari2016xnor, geng2019lp, ar119}. It is anticipated that many industrial companies follow the scaling roadmap shown in Fig. \ref{fig:BNN_trend} and expeditiously adopt the reduced precision representations in Deep NN (DNN) training and inference.

However, the precision reduction in the representative data has exacerbated the impact of soft errors in the quantized neural network (QNN). In particular, compressing the information of a 32-bit data into few bits increases the risk of losing a significant portion of the tensors' information, which are constructed by reduced precision parameters. On the other hand, the current trend of aggressive dimension reduction in the transistors has intensified this reliability challenge.  The soft errors are generally originated from high-energy particles and might strike the sequential/combinational logic in the ML/DL accelerators. The DL accelerators systematically perform the inference operation in a pre-trained network. Thus, it is expected that the neural network (NN) inference accelerator to maintain its functionality for an extended period. However, the accumulated radiation-induced transient faults can potentially impact the individual parameters in NN topology, and if not mitigated immediately, the functionality of a long-running expected NN inference accelerator can gradually downgrade 
by outlier contamination \cite{Sedghi2019MLSP}, and lead to drastic accuracy loss.

Based on our empirical observations obtained from the rigorous NN vulnerability assessment against soft errors, we observed that the weights and activation functions are unevenly susceptible to both single-event upset (SEU), and multi-bit upset (MBU) scenarios. We also noticed that not only the soft errors inconsistently influence the parameters, but the magnitude of fault-influence on the accuracy loss is subject to how early the parameters in the consecutive layers are exposed to soft errors.

In this paper, we propose an entropy-based model to estimate the impact of the soft errors on the NN architectural model. Our proposed model delineates the effects of both undertaken SEU and MBU  across layers and per each layer of the selected NN. The proposed model can be utilized by NN architectural developers to evaluate the error-resiliency magnitude of the NN before employing it in the safety-critical applications. In summary, our significant contributions in this paper can be listed as follows:

1) A comprehensive fault injection methodology is used to delineate the fault-skeleton map which reveals the sensitivity of each NN parameter in each layer of NN to the soft errors. 

2) Based on the results collected in the previous step, we propose an entropy-based model to estimate the impact of the soft errors on the NN architectural. This model stresses on how the appearance of fault at different sequence of layers and parameters can potentially lead to drastic misclassification.

3) Our finding highlights that considering equal importance for the parameters across the layers is inaccurate, an assumption that does not hold according to our preliminary assessment of the layer importance. Our findings suggest then to consolidate traditional assumption by incorporating our preliminary finding for layer importance. 
\\
The remainder of the paper is organized as follows. Section II presents the background on convolutional NN. Section III describes the empirical observations on fault injection in the NN accelerator. Section IV explains our entropy-based model for estimating soft error impact in NN topology. In Section V, the experimental results are presented. Finally, Section VI concludes the paper.




\section{Background}
\begin{figure}{t!}
  \vspace{-25pt}
  \begin{center}
    \includegraphics[width=0.49\textwidth]{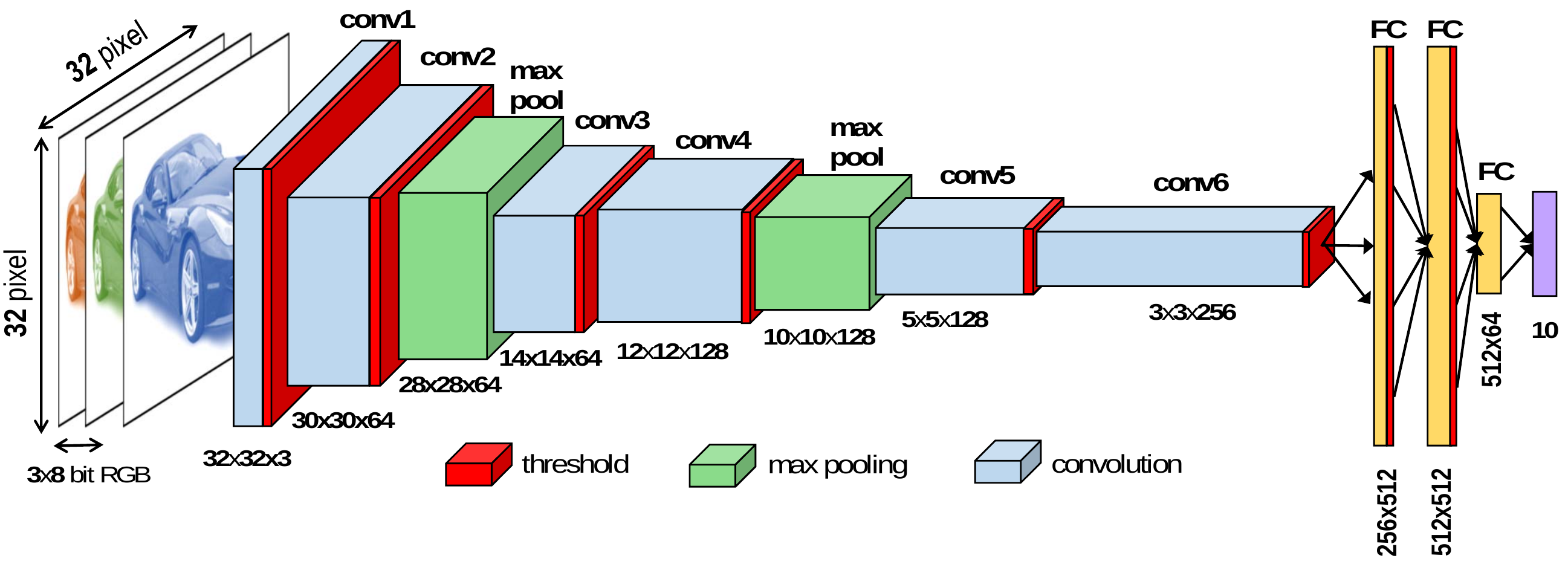}
  \end{center}
 \vspace{-18pt}
  \caption{a schematic view of cnvW2A2 architecture designed in FINN \cite{umuroglu2017finn}.}
    \label{fig:cnv}
\end{figure}

The Convolutional Neural Network (CNN) is generally used to extract the features of unique information through a hierarchy of layers. The CNN is widely applied in a variety of applications such as image processing \cite{chen2017deeplab}, sentence classification \cite{ kim2014convolutional}, semantic parsing \cite{yih2015semantic}, and speech recognition \cite{abdel2014convolutional}. As illustrated in Fig. \ref{fig:cnv}, CNN is constructed by different types of layers including \textit{convolutional}, \textit{fully connected}, and \textit{pooling} layers. The input layer provides a set of information based on the applications of CNN. For instance, in order to classify the image, the input layer retrieves raw pixel values from the image \cite{CNN_note}. Furthermore, the \textbf{\textit{CNN layers}} are explicitly described as the following four sub-layers:
\\
1) \textit{convolution sub-layer}: A dot product operation is performed among the weights of the regionally-connected neurons and their input sets to compute the output, 
\\
2) \textit{non-linear sub-layer}: It employs an activation function to map the weight sum of regionally-connected neurons to $max(0,$ \textit{weight sum of regionally-connected neurons}$)$, 
\\
3) \textit{normalization sub-layer}:  It performs the scaling of the feature values   \cite{stackoverflow}, 
\\
4) \textit{pool sub-layer}: It reduces the spatial size to decrease the number of parameters and computations within the network \cite{CNN_note}. 
Lastly, the \textit{Fully Connected (FC) layer} takes the previous \textit{CNN layers} and transforms it as a vector that represents the set of feature values.

\section{Empirical Observations on Fault Injection on NN Accelerator}

\subsection{Fault Injection Category}
Regardless of how dynamic the soft errors are, the functionality of both combinational and sequential logic circuits can be either partially or entirely sabotaged upon the strike of radiation-induced transient faults. The effect of soft errors is generally modeled as a uniform distribution across space and time. This assumption is in line with the study presented in \cite{dixit2011impact}. To precisely examine the impact of soft errors on the NN accelerator, both Single-Event Upset (SEU) and Multi-Bit Upset (MBU) are studied in our rigorous NN evaluation. Even though the SEUs are considered as the significant source of transient faults, the study in \cite{dixit2011impact} shows that the fragment of systems that are impacted by MBUs have increased over the past years. The primary driver of this movement is the aggressive transistor dimension reduction, which enables the integration of low dimension transistors in an ultra small scale. This aggressive accommodation enables the radiation-induced transient faults with less energy to unbalance the critical charge of adjacent transistors.  To mimic the behavior of an MBU, we followed the recommendation in \cite{schirmeier2016efficient} by altering a burst of bits in size of 8-bit. For the sake of simplicity, we disregard the faults that might occur in CPU, main memory, memory bus, and the combinational logic.

\subsection{Fault Injection on NN Parameters}
The parameters that are used for studying the SEU/MBU impact are described as follows:

\begin{itemize}
    \item 
    \textbf{\textit{Weights}}: The weights in the NN accelerator remain untouched during the inference operation. However, if a weight tensor is contaminated by an error that stems from soft errors, it has the potential to contaminate the consecutive layers in NN.

    \item 
    \textbf{\textit{Activations}}:
    The activation tensor of each convolution layer is used as the input for the convolution operation in the next layer. If the soft error strikes on regionally-connected activation nodes, the accelerator could experience certain degree of accuracy loss.  
  
    \item 
    \textbf{\textit{Layers}}: The convolutional network topology is generally composed of multiple convolutional layers, max pool layers, and fully connected layers. Since each category of layers has distinct parameters, they can contribute non-uniformly to the accuracy loss in the NN accelerator.
    
\end{itemize}

\subsection{Fault Injection Distribution}
\noindent\textbf{Uniform fault injection across layers:} The SEU/MBU can be uniformly injected to the NN architecture across the space and time of memory dimension. Besides, depending on the points of fault injection, soft errors can be inserted into the given memory location where the particular parameters are stored and executed for the NN topology. 

\noindent\textbf{Targeted in-layer fault injection:} Considering that the convolutional layers are usually customized with various tensor dimension sizes, the in-layer faults can make an impact on the classification accuracy. Such an in-layer fault can represent itself as either a dramatic accuracy drop or reasonable accuracy loss/gain. Our study will present the impact of soft errors against targeted parameters within the layer.

\subsection{Experimental Results Analysis}
\begin{figure}[t!]
\centering
    \includegraphics[width=0.48\textwidth]{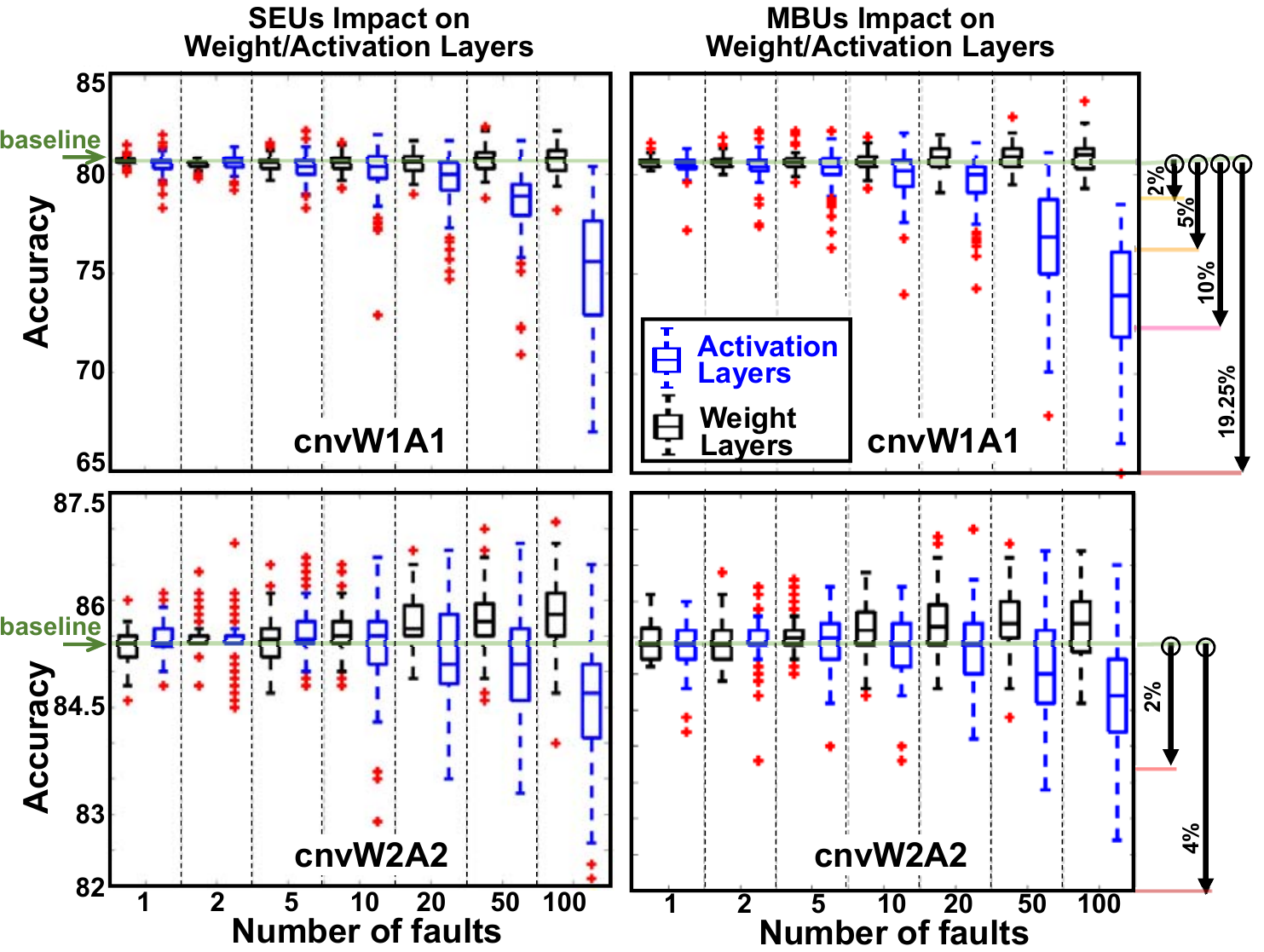}
    \vspace{-10pt}
  \caption{The impact of uniform 
  SEUs/MBUs injection on two individual groups of layers: (i) the stack of weight layers, and (ii) the stack of activation layers, across the entire network in both \underline{cnvW1A1} and \underline{cnvW2A2} networks.}
    \label{fig:fault_dis_before_TMR}
  \vspace{-10pt}
\end{figure}

\begin{figure}[t!]
\centering
    \includegraphics[width=0.5\textwidth, height=8cm]{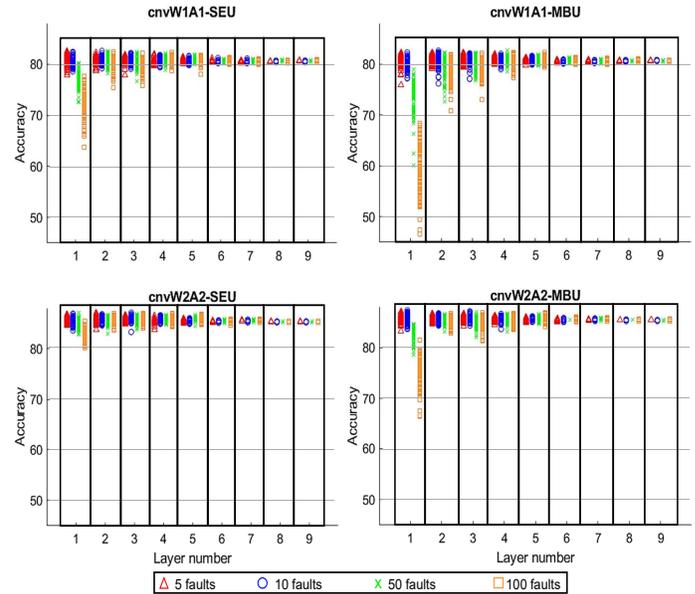}
  \vspace{-23pt}
  \caption{The impact of targeted in-layer 
  SEUs/MBUs injection on both \underline{unprotected cnvW1A1} and \underline{unprotected cnvW2A2} networks.}
    \label{fig:in_layer}
\end{figure}

The sensitivity of the NN accelerator to both SEU/MBU across the stack of layers in NN and per each individual layer of NN are evaluated. Based on our experimental results demonstrated in Fig. \ref{fig:fault_dis_before_TMR} and Fig. \ref{fig:in_layer}, we observed the following insights:
\begin{itemize}
    \item Reducing the number of bits for storing the network information increases the vulnerability of the NN accelerator to soft errors (SEU/MBU). 
    \item The activation layers are considerably vulnerable to both SEU and MBU, as shown in Fig. \ref{fig:in_layer}.
    \item The classification accuracy is gradually dropped with respect to the number of accumulated SEU/MBU.
    \item The effect of MBU is relatively higher than SEU upon the NN accelerator.
    \item The NN accelerator can still suffer from a \underline{drastic} misclassification in the worst-case scenarios despite the negligible average degradation due to the soft errors' effect. 
    For instance, the accuracy of image classifier can drastically drop by \textbf{19.25\%} in \textit{cnvW1A1} where 100 MBU events are injected during the workload operation as illustrated in Fig. \ref{fig:fault_dis_before_TMR}. 
\end{itemize}


\section{Building Model Inspired by Empirical Observations}
We have recently discovered the existence of worst-case scenarios exists with drastic degradation on the network inference accuracy. To understand and design mitigation techniques, and to delineate the impact of both undertaken SEU and MBU across layers and per each layer of the selected NN, we, therefore, performed a statistical analysis of our empirical database explained here. To stand on a firm basis, let us first review the basics of our database and model developments. Vulnerability, in our context, defines the level to which a bit under the fault injection may perturb inference accuracy. Accuracy is, therefore, defined as the distance between the network inferred output and the actual output. The distance also referred to as loss, is formulated as below,
\begin{equation}
    \calL(\rmf(x;(w_l)_{l=1}^\frakL,s)
\end{equation}
where $x$ and $s$ are the input and target output for the trained binarized network with $\frakL$ convolutional layers. The $\rmf(.)$ represents network inference computation, and $\calL(.)$ is the computed loss between the network output and target output. It is worth to mention that weights are often stored in their two's complement after being converted to binarized form. The quantization, into 2-bits, is achieved through either deterministic,
\begin{equation}
    w =
    \begin{cases}
      +1 & \text{if $w^r \geq 0$}\\
      -1 & \text{otherwise}\\
    \end{cases}       
\end{equation}
or stochastic approach,
\begin{equation}
    w =
    \begin{cases}
      +1 & \text{with probability of $\sigma(w^r)$}\\
      -1 & \text{with probability of 1-$\sigma(w^r)$}\\
    \end{cases}       
\end{equation}
\begin{equation}
    \sigma(w) = max(0, min(1,\frac{w^r+1}{2})) 
\end{equation}
where $w^r$ is the real-world calculated weight before quantization. The stochastic approach is more appealing, while the choice depends upon the computational constrained. During the training phase, the objective is to minimize the amount of loss. Without the loss of generality, the training process can readily be translated into the form of an optimization problem,

\begin{equation}
    min_{w_l}(\calL(\rmf(x;(w_l)_{l=1}^\frakL,s))
\end{equation}
From our modeling perspective, the optimization problem is reversely seen where the objective is to inject faults in the forms of bit-flips, such that the resulting network operation is perturbed,
\begin{equation}
    max_{\tilw}(\calL(\rmf(x;(\Tilde{w_l})_{l=1}^\frakL,s) - \calL(\rmf(x;(w_l)_{l=1}^\frakL,s))
\end{equation}
where $\Tilde{w}$ represents perturbed weights due to the fault injection. It is worth to emphasize that in the bit-flip attack (BFA) model \cite{rakin2019bit,courbariaux2016binarized}, the distance between pre-fault and post-fault weight tensors shall remain below a preset. Targeted BFA implementation is, therefore, transcribed as a bit search algorithm to solve the above-mentioned optimization problem constrained on finding an optimal combination of vulnerable bits to perturb.\\

\subsection{Modeling the Impact of Uniform fault injection across layers}
We not only analyzed the acquired results to determine the most vulnerable NN parameters to the soft error but also to objectively build the corresponding statistical model that describes the weighted impact of fault-contaminated parameters on the inference accuracy of NN. 
The designed experiment included injection of SEU- and MBU-based uniform faults over \textit{cnvW1A1} and \textit{cnvW2A2} network architectures. The experiment was aimed on surveying the impact of four input variables, namely (1) quantization level for the network architecture under study ($X_{1} \in [1,2]$), (2) fault mode ($X_{2} \in {SEU, MBU}$), (3) identification variable for the fault domain ($X_{3} \in {weight, activation}$), (4) number of faults ($X_{4} \in [1,2,5,10,20,50,100]$), on the inferred accuracy, as hypothesized by,
\begin{equation}
\begin{split}
   E[Y_{t}|X] = \beta_{0} + \beta_{1}X_{t1} + \beta_{2}X_{t2} + \beta_{3}X_{t3} + \beta_{4}X_{t4} + \\
   \beta_{12}X_{t1}X_{t2} + \beta_{13}X_{t1}X_{t3} + \beta_{14}X_{t1}X_{t4} + \beta_{23}X_{t2}X_{t3} +\\ \beta_{24}X_{t2}X_{t4} + \beta_{123}X_{t1}X_{t2}X_{t3} + \beta_{234}X_{t2}X_{t3}X_{t4} +\\ \beta_{1234}X_{t1}X_{t2}X_{t3}X_{t4}+\epsilon
\end{split}
\end{equation}
where $\beta{1}$, $\beta{2}$, $\beta{3}$, $\beta{4}$, $\beta{12}$, $\beta{13}$, $\beta{14}$, $\beta{23}$, $\beta{24}$, $\beta{123}$, $\beta{234}$, $\beta{1234}$ define the set of coefficient parameters ($B$), for the set of variables, previously defined. The model coefficients set $B$ were then estimated by,
\begin{equation}
 \hat{B}\equiv \arg\min_{B} (\sum_{t} (Y_{t} - \hat{Y_{t}})^2) 
\end{equation}

Table \ref{tab:cnv} summarizes the calculated set of coefficients ($B$), along with their corresponding standard error and statistical significance level, for the derived model. Our derived model resulted in an adjusted $R^2$ of 0.91 ($p<0.01$). We found a significant interaction among all independent variables for the inferred accuracy prediction.

\begin{table}[b!]
    \centering
        \caption{Estimates of fault model coefficients for SEU- and MBU- based fault. Coefficients with significance level of $<.05$ and $<.01$ are respectively tailed by * and ** symbols.}
        \vspace{-10pt}
    \fontsize{7}{7.2}\selectfont
    \begin{tabular}{*4c}
    \toprule
     parameter & estimate & std. error & $t$-value\\
    \midrule
    \hline
    $\beta_{0}$ & 74.76** & 0.68 & 110.48 \\
    $\beta_{1}$ & 5.37** & 0.43 & 12.49 \\
    $\beta_{2}$ & 0.45 & 0.42 & 1.07 \\
    $\beta_{3}$ & 0.65 & 0.41 & 1.58 \\
    $\beta_{4}$ & 0.06** & 0.01 & 4.18 \\
    $\beta_{12}$ & -0.19 & 0.27 & -0.7 \\
    $\beta_{13}$ & -0.33 & 0.27 & -0.7 \\
    $\beta_{14}$ & -0.02* & 0.01 & -2.27 \\
    $\beta_{23}$ & -0.28 & 0.26 & 1.11 \\
    $\beta_{24}$ & 0.04** & 0.01 & 4.68 \\
    $\beta_{34}$ & -0.06** & 0.01 & -6.8 \\
    $\beta_{123}$ & 0.12 & 0.16 & 0.74 \\
    $\beta_{124}$ & -0.02** & 0.00 & -4.00 \\
    $\beta_{134}$ & 0.02** & 0.00 & 4.09 \\
    $\beta_{234}$ & -0.04** & 0.00 & -7.13 \\
    $\beta_{1234}$ & 0.02** & 0.00 & 5.93 \\
    \hline
    \end{tabular}
    \label{tab:cnv}
\end{table}{}

\begin{figure}[t!]
\centering
    \includegraphics[width=0.5\textwidth]{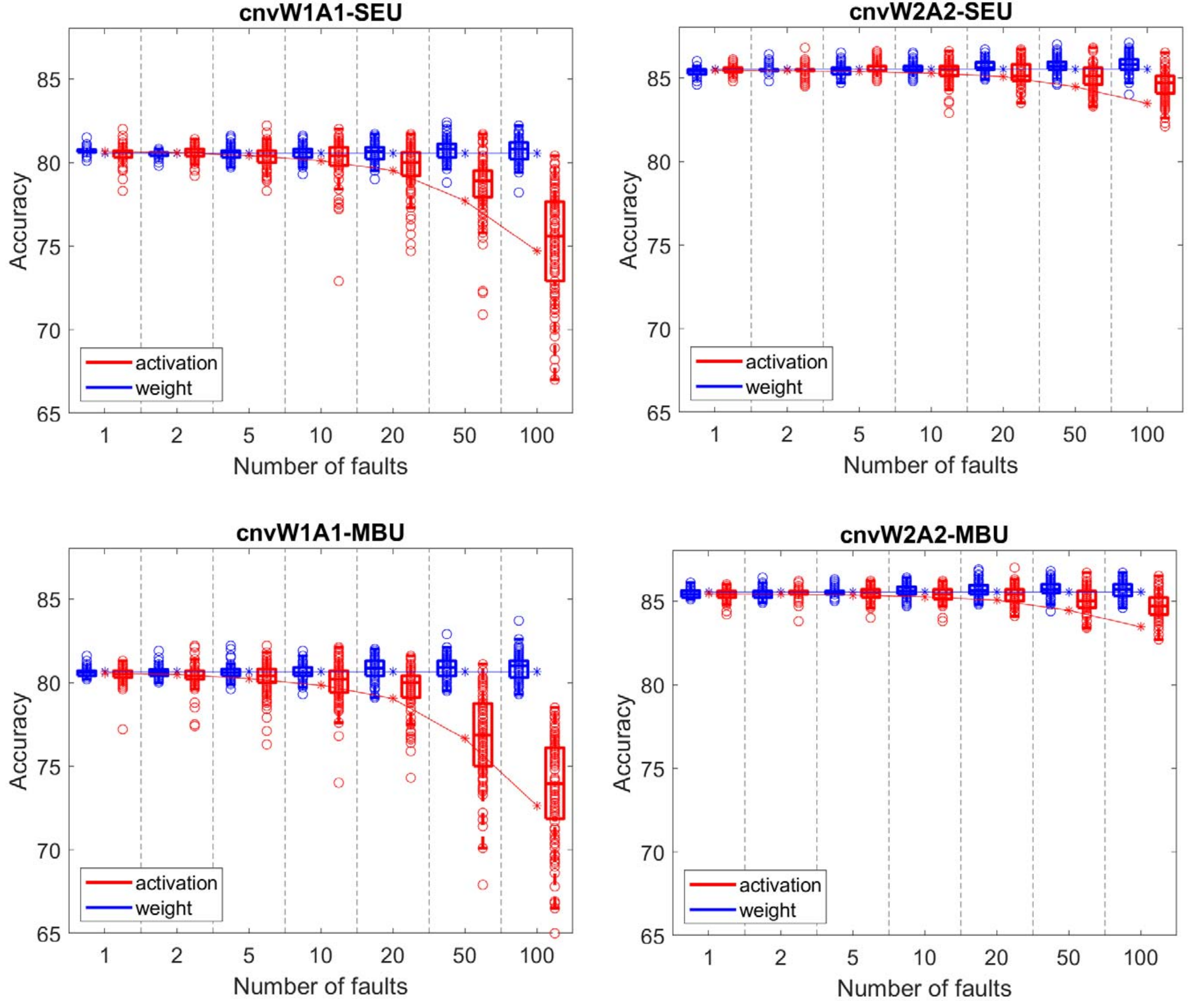}
  \vspace{-10pt}
  \caption{The impact of uniform 
  SEUs/MBUs injection on two individual groups of layers: (i) the stack of weight layers, and (ii) the stack of activation layers, across the entire network in both \underline{unprotected cnvW1A1} and \underline{unprotected cnvW2A2} networks. The impacts are overlaid with the predicted.}
    \label{fig:across_NN}
\end{figure}

\subsection{Modeling the Impact of Targeted In-layer fault injection}
\begin{figure*}[t!]
\centering
    \includegraphics[width=1\textwidth]{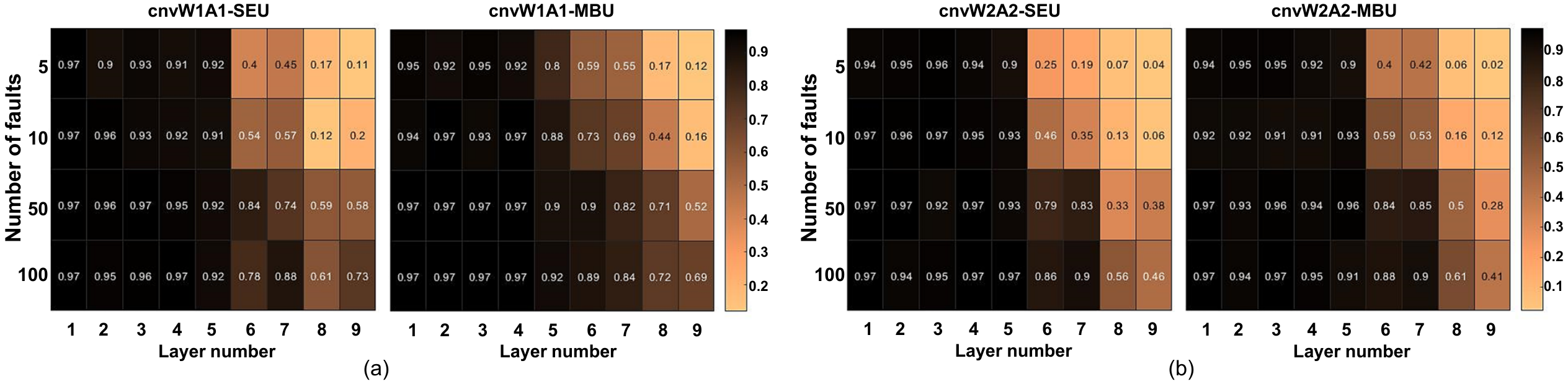}
  \vspace{-10pt}
  \caption{The probability of accuracy drop is written in each square based on the number of accumulated SEU/MBU injection per each layer in (a) \underline{unprotected cnvW1A1} and (b) \underline{unprotected cnvW2A2}.}
    \label{fig:heatMap}
\end{figure*}

To evaluate the importance of layer number on fault modeling, we conducted a separate experiment (injection of SEU- and MBU-based uniform faults over \textit{cnvW1A1} and \textit{cnvW2A2} network architectures),  with the following independent variables: (1) quantization level for the network architecture under study ($X_{1} \in [1,2]$), (2) fault mode ($X_{2} \in {SEU, MBU}$), (3) the layer number in which the fault was injected ($X_{1} \in[1,2, ..., 9]$),  and (4) number of faults ($X_{4} \in [5,10,50,100]$), on the inferred accuracy, as hypothesized by,
\begin{equation}
\begin{split}
   E[Y_{t}|X] = \beta_{0} + \beta_{1}X_{t1} + \beta_{2}X_{t2} + \beta_{3}X_{t3} + \beta_{4}X_{t4} + \\
   \beta_{12}X_{t1}X_{t2} + \beta_{13}X_{t1}X_{t3} + \beta_{14}X_{t1}X_{t4} + \beta_{23}X_{t2}X_{t3} +\\ \beta_{24}X_{t2}X_{t4} + \beta_{123}X_{t1}X_{t2}X_{t3} + \beta_{234}X_{t2}X_{t3}X_{t4} +\\ \beta_{1234}X_{t1}X_{t2}X_{t3}X_{t4}+\epsilon
\end{split}
\end{equation}
where $\beta{1}$, $\beta{2}$, $\beta{3}$, $\beta{4}$, $\beta{12}$, $\beta{13}$, $\beta{14}$, $\beta{23}$, $\beta{24}$, $\beta{123}$, $\beta{234}$, $\beta{1234}$ define the set of coefficient parameters ($B$), for the set of variables, previously defined. The model coefficients set $B$ were then estimated by,
\begin{equation}
 \hat{B}\equiv \arg\min_{B} (\sum_{t} (Y_{t} - \hat{Y_{t}})^2) 
\end{equation}
Table \ref{tab:cnvLayers} summarizes the found coefficients. This finding highlights that considering equal importance for the parameters across the layers is inaccurate, an assumption that does not hold according to our preliminary assessment of layer importance. Adjusted $R^2$ was found to be 0.69 ($p<0.01$). We hypothesize that the reduction in the value of $R^2$ is due to the non-existence of a model variable concerning the fault domain in this model. Considering the observed non-uniform behavior across layers, we surveyed an entropy-based measure to estimate a layer likelihood to be fault-tolerant. Heat-maps in Fig. \ref{fig:heatMap} display the probability distributions of the estimated likelihood of each layer being fault-tolerant (in 100 trials) over different architectures and under different fault modeling condition. We observe that the first layers are subjected to more sensitivity than the last layers. This finding highlights that considering equal importance for the parameters across the layers is inaccurate, an assumption that does not hold according to our preliminary assessment of layer importance. Our findings suggest then to consolidate traditional assumption by incorporating our preliminary finding for layer importance. We refer to this approach as the entropy-based modeling technique, a technique where our a priori information about the spatial importance is utilized to derive proper design of the computational networks.

\begin{table}[b!]
    \centering
        \caption{Estimates of fault model coefficients for SEU- and MBU- based fault. Coefficients with significance level of $<.05$ and $<.01$ are respectively tailed by * and ** symbols.}
        \vspace{-10pt}
    \fontsize{7}{7.2}\selectfont
    \begin{tabular}{*4c}
    \toprule
     parameter & estimate & std. error & $t$-value\\
    \midrule
    \hline
    $\beta_{0}$ & 75.26** & 0.65 & 116.95 \\
    $\beta_{1}$ & 5.10** & 0.41 & 12.33 \\
    $\beta_{2}$ & 0.79 & 0.41 & 1.95 \\
    $\beta_{3}$ & 0.08 & 0.14 & 0.5 \\
    $\beta_{4}$ & 0.06** & 0.01 & 5.56 \\
    $\beta_{12}$ & -0.17 & 0.26 & -0.67 \\
    $\beta_{13}$ & -0.04 & 0.09 & -0.42 \\
    $\beta_{14}$ & -0.01* & 0.01 & -2.23 \\
    $\beta_{23}$ & -0.12 & 0.09 & -1.38 \\
    $\beta_{24}$ & -0.11** & 0.01 & -17.33 \\
    $\beta_{34}$ & -0.01** & 0.00 & -4.07 \\
    $\beta_{123}$ & 0.03 & 0.06 & 0.44 \\
    $\beta_{124}$ & 0.03** & 0.00 & 8.29 \\
    $\beta_{134}$ & 0.00 & 0.00 & 1.64 \\
    $\beta_{234}$ & 0.01** & 0.00 & 12.75 \\
    $\beta_{1234}$ & 0.00** & 0.00 & -5.90 \\
    \hline
    \end{tabular}
    \label{tab:cnvLayers}
\end{table}{}

\section{Experimental Setup}

\noindent\underline{\textbf{Dataset:}} 
In our experiments, we consider the convolutional network topology (\textbf{CNV}) inspired by BinaryNet \cite{courbariaux2016binarized} and VGG-16 \cite{VGG_16}. It is tailored with six convolutional layers, three max pool layers, and three fully-connected layers. The classification process for this dataset utilizes a CNN to classify images from the CIFAR 10 image set. Two CNNs were tested: one using 1-bit weight and activations, and one using 2-bit weights and activations.  There are around \underline{1.6 million} susceptible bits and \underline{3.2 million} susceptible bits to soft errors in 1-bit and 2-bit CNVs, respectively.

\noindent\underline{\textbf{Experiment Setup:}} 
We extensively modified the BNN-PYNQ project to perform 2000 fault injections for each scenario to collect a sufficient pool of tests. This pool relatively presents a comprehensive sample of the behavior of fault in each scenario assumption.  The faults are injected on the targeted parameters either across the entire NN or per a specific layer.  We limit our experimental results on classifying 1000 images in CIFAR-10 to reduce the relative long fault injection process while still delivering an approximately-sufficient representation of the original consideration. 
The wights and activations are considerably reused in the course of inference operation. They are stored on on-chip buffers to reduce the memory access time significantly. We evaluated the sensitivity of the NN accelerator to a range number of faults that might occur in the operational lifetime of the device. The faults are injected at uniformly distributed times either across the entire NN or per each individual layer during the image classification. A host program running on the ARM processor initializes the weights and activations of the network on the FPGA, then prepares the image set and launches the classification process. During the image classification, faults are injected by reading from memory on the FPGA used for CNN’s parameters, flipping a bit or a word, then writing back the result. The outputs of the image classification are then compared against the correct labels to calculate accuracy.

\section{Conclusions}
The soft errors can potentially change the content of each individual parameter in \textit{NN} topology, and if not mitigated immediately, the accumulated faults can gradually downgrade the functionality of a long-running expected \textit{NN} inference accelerator. This might lead to drastic image misclassification that can be considered as a delicate reliability concern in safety-critical applications. In this paper, we proposed an entropy-based model to estimate the impact of the soft errors on the \textit{NN} architectural. Our proposed model delineates the impact of both undertaken SEU and MBU  across layers and per each layer of the selected \textit{NN}. It can be utilized by \textit{NN} architectural developers to evaluate the error-resiliency magnitude of the \textit{NN} before employing it in the safety-critical applications.

\bibliographystyle{IEEEtran}
\fontsize{8}{7.2}\selectfont
\bibliography{biblio}
\end{document}